\documentclass[letterpaper]{article} 
\usepackage[preprint]{aaai2027}  
\usepackage[hyphens]{url}  
\usepackage{graphicx} 
\usepackage{natbib}  
\usepackage{caption} 
\usepackage{algorithm}
\usepackage{algorithmic}

\usepackage{newfloat}
\usepackage{listings}
\DeclareCaptionStyle{ruled}{labelfont=normalfont,labelsep=colon,strut=off} 
\floatstyle{ruled}
\newfloat{listing}{tb}{lst}{}
\floatname{listing}{Listing}

\usepackage{booktabs}

\usepackage{amsmath}
\usepackage{amssymb}  

\title{
PBD-AG:
Persistent Baseline-Delta Active Graphs with Uncertainty-Aware Inspection for Long-Horizon Service Robots
}
\author{
Shuo Bao\textsuperscript{\rm 1,\rm 8}\equalcontrib,
Wei Dong\textsuperscript{\rm 4,\rm 8}\equalcontrib,
Shuyue Zhang\textsuperscript{\rm 6,\rm 8},
Ming Shang\textsuperscript{\rm 3,\rm 8},\\
Yuchen Huang\textsuperscript{\rm 7,\rm 8},
Han Yu\textsuperscript{\rm 4,\rm 8},
Chengjie Xu\textsuperscript{\rm 3,\rm 8},
Yiheng Bi\textsuperscript{\rm 5,\rm 8},\\
Kai Sun\textsuperscript{\rm 2,\rm 8},
Fuchun Sun\textsuperscript{\rm 2,\rm 8}\corresponding,
Xinzhou Wang\textsuperscript{\rm 2,\rm 8}\corresponding
}
\affiliations{
\textsuperscript{\rm 1}Peking University, Beijing, China\\
\textsuperscript{\rm 2}Tsinghua University, Beijing, China\\
\textsuperscript{\rm 3}Beihang University, Beijing, China\\
\textsuperscript{\rm 4}Beijing Institute of Technology, Beijing, China\\
\textsuperscript{\rm 5}Yale University, New Haven, Connecticut, USA\\
\textsuperscript{\rm 6}Beijing University of Chemical Technology, Beijing, China\\
\textsuperscript{\rm 7}University of Science and Technology Beijing, Beijing, China\\
\textsuperscript{\rm 8}Forge Robotics, Beijing, China\\
baoshuo26@stu.pku.edu.cn, 3220240638@bit.edu.cn,\\
fcsun@tsinghua.edu.cn, 709510112@qq.com
}

\begin{document}
\maketitle

\begin{abstract}
Long-horizon service robots require persistent world models that can be built autonomously in unseen environments and revised as task-relevant objects change. Existing methods rely on online mapping, which accumulates localization and observation errors, static scene representations that cannot capture persistent object changes, or holistic vision-language predictions that lack verifiable 3D geometric evidence. We present PBD-AG, a persistent baseline-delta active graph framework that decouples robot-verified stable fixtures from revisable dynamic object events. Under our framework, the robot autonomously bootstraps the structural baseline from onboard exploration and inspects discovered fixtures to ground hierarchical object beliefs. PBD-AG maintains reliability-weighted object states over geometry, semantics, identity, existence, and support relations, utilizing a geometric visibility gate to mitigate false deletions under occlusion. Inspection viewpoints are selected by a graph-conditioned policy that balances target coverage, travel cost, collision risk, and redundant observation. Simulation experiments in multiple environments and under controlled dynamic evaluation show higher aggregate coarse-fixture F1 than capability-matched controls, as well as stronger identity continuity and event recall. A qualitative physical-robot demonstration further illustrates integration with onboard sensing, providing a traceable world model for long-horizon robotic perception.
\end{abstract}

\begin{figure}[t]
    \centering
    \includegraphics[width=\columnwidth]{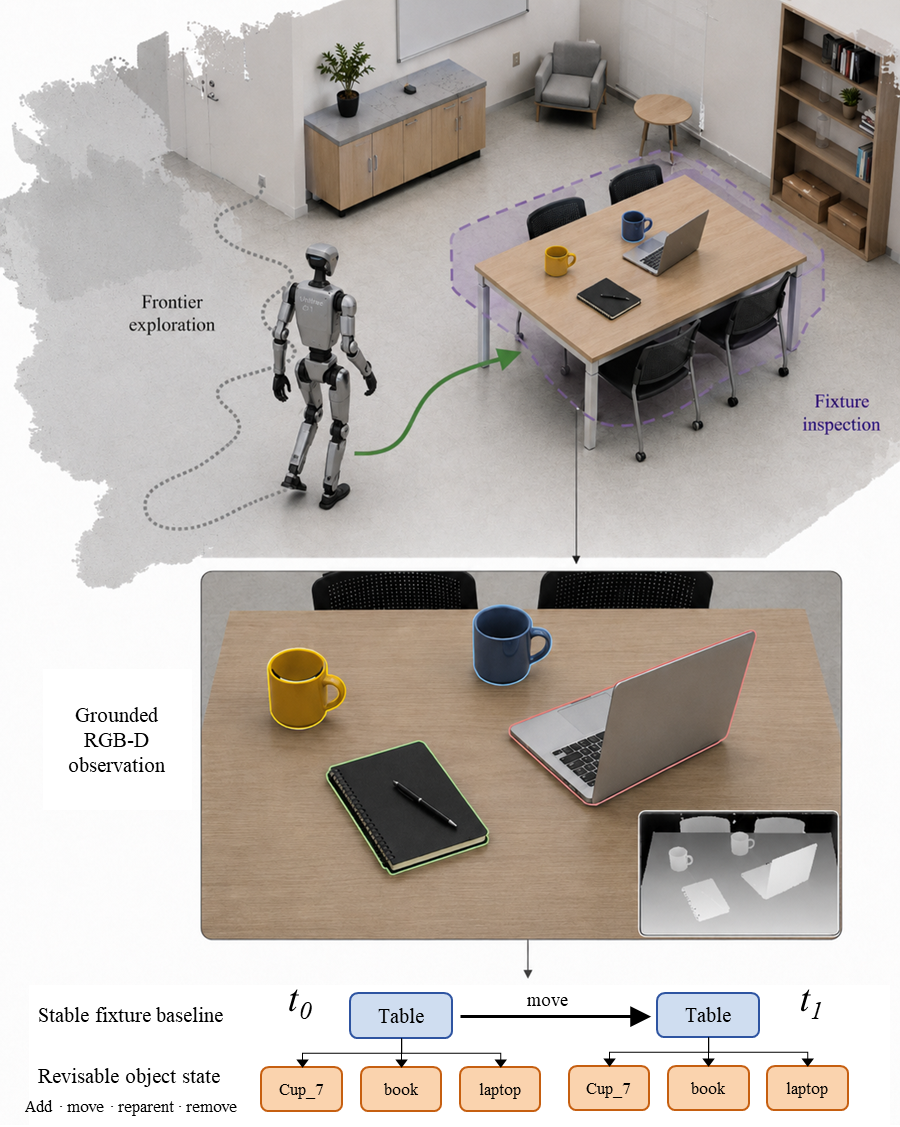}
    \caption{The robot first discovers stable fixtures through exploration and later maintains fine-grained object changes through event-based graph updates.}
    \label{fig:teaser}
\end{figure}

\section{Introduction}

Long-horizon service robots operate in environments whose structural layout changes slowly, while task-relevant objects may appear, disappear, move, or transfer between support surfaces. A useful world model must therefore support reliable state revision from onboard observations, rather than merely reconstructing the scene at a single time. Static maps cannot represent such changes, whereas rebuilding the entire semantic map after every revisit is inefficient and prone to identity fragmentation.

Open-vocabulary maps provide flexible language grounding
\citep{huang2023vlmaps,jatavallabhula2023conceptfusion}, while object-centric
scene graphs offer compact relational abstractions
\citep{gu2024conceptgraphs,werby2024hovsg}. However, these approaches are
primarily evaluated as static reconstructions. Dynamic memories support online
state updates \citep{liu2024dynamem,yan2025dovsg}, yet a missed detection alone
cannot distinguish physical removal from occlusion, poor viewpoint, or detector
failure. Treating repeated misses as evidence of absence can delete an object
that remains present, whereas processing each observation independently can
create duplicate identities for a moved object. These failure modes become
consequential when downstream planners rely on object identities and support
relations across sessions.

To address these challenges, we introduce PBD-AG, a persistent
baseline--delta active graph framework for long-horizon scene understanding.
Its key premise is to separate scene information according to its rate of
change and the evidence required to revise it. PBD-AG begins with no preloaded
fixture graph. Frontier exploration expands the occupancy map, accumulated
RGB-D evidence generates structural fixture hypotheses, and reachable
fixture-centered observations verify their semantics and geometry. Repeatedly
verified fixtures form a versioned baseline. Fine-grained objects are
associated with these fixtures, and their subsequent changes are maintained
as typed event deltas rather than through repeated global reconstruction
(Figure~\ref{fig:teaser}).

Each persistent node integrates semantic, geometric, appearance, existence,
and support-parent evidence. A missed detection contributes negative evidence
only when the predicted 3D extent lies within the camera frustum, is not
occluded by closer observed geometry, and has sufficient expected pixel
support. The graph places unconfirmed or conflicting fixture states in an
inspection queue, while global spatial coverage remains governed by standard
frontier exploration rather than semantic predictions over unseen space.
Our contributions are:
\begin{itemize}
    \item We propose a robot-bootstrapped baseline--delta graph that learns
stable fixtures through onboard exploration and represents subsequent
object changes as typed event deltas in a shared metric frame, without
requiring a preloaded scene-specific semantic graph.

    \item We introduce a persistent object state that integrates multiple
evidence sources, with a geometric visibility gate that conditions
negative evidence on object observability to maintain existence, identity,
geometry, and support relations over time.

    \item We use the persistent graph to drive active inspection, interleaving
standard frontier exploration with reachable fixture-centered observations.
Candidate views balance target coverage against travel cost, collision risk,
observation redundancy, and model-call limits.

    \item We design three controlled evaluation protocols that separately
isolate representation, temporal-memory, and active-acquisition effects:
shared fixed-trajectory replay, a shared-observation dynamic benchmark,
and paired independent active rollouts.
\end{itemize}
We also qualitatively demonstrate the complete perception-to-graph pipeline on a physical robot using onboard RGB-D sensing.

\section{Related Work}

\paragraph{Open-vocabulary metric maps and scene graphs.}
VLMaps and ConceptFusion fuse language-aligned features into metric maps
\citep{huang2023vlmaps,jatavallabhula2023conceptfusion}. ConceptGraphs
associates multi-view object proposals into an open-vocabulary graph
\citep{gu2024conceptgraphs}, while HOV-SG organizes floors, rooms, and objects
hierarchically \citep{werby2024hovsg}. Hydra incrementally constructs and
optimizes metric-semantic scene graphs \citep{hughes2022hydra}; Clio selects
task-relevant graph granularity online \citep{maggio2024clio}. These systems primarily target scene reconstruction or semantic querying. These systems do not explicitly address how to distinguish a persistent
real-world change from an unreliable observation.

\paragraph{Dynamic robot memory.}
DynaMem builds a mutable spatio-semantic voxel memory online from posed
RGB-D observations and can explore previously unseen environments
\citep{liu2024dynamem}. However, its point-centric representation does not explicitly maintain
persistent object identities and hierarchical support relations, nor does it record object changes as typed events. DovSG constructs an
open-vocabulary 3D scene graph from an initial RGB-D scan and subsequently
updates affected subgraphs during long-term manipulation
\citep{liu2024dynamem,yan2025dovsg}. Its dynamic-update stage therefore begins from an
initialized global scene graph. In contrast, PBD-AG couples frontier
exploration, fixture discovery, and reachable close-view confirmation to
bootstrap a structural baseline from an empty scene-specific graph. It then
maintains object-level identity, existence, and support-parent beliefs through
typed event deltas, admitting negative evidence only after geometrically valid
repeated misses.

\paragraph{Active exploration and inspection.}
Frontier exploration selects boundaries between known free and unknown space
\citep{yamauchi1997frontier}; learned exploration policies additionally exploit
geometric regularities \citep{chaplot2020active}. VLFM assigns language-based
values to frontiers for zero-shot object navigation
\citep{yokoyama2024vlfm}, and SCOUT couples semantic uncertainty to scene
coverage \citep{mao2026scout}. PBD-AG instead uses conventional frontiers for
unknown-space coverage and persistent graph state to trigger close observation
of already discovered fixtures. This separation avoids attributing semantic
content to unexplored regions while still directing sensing toward unresolved
relational state.

\begin{figure*}[t]
    \centering
    \includegraphics[width=0.90\textwidth]{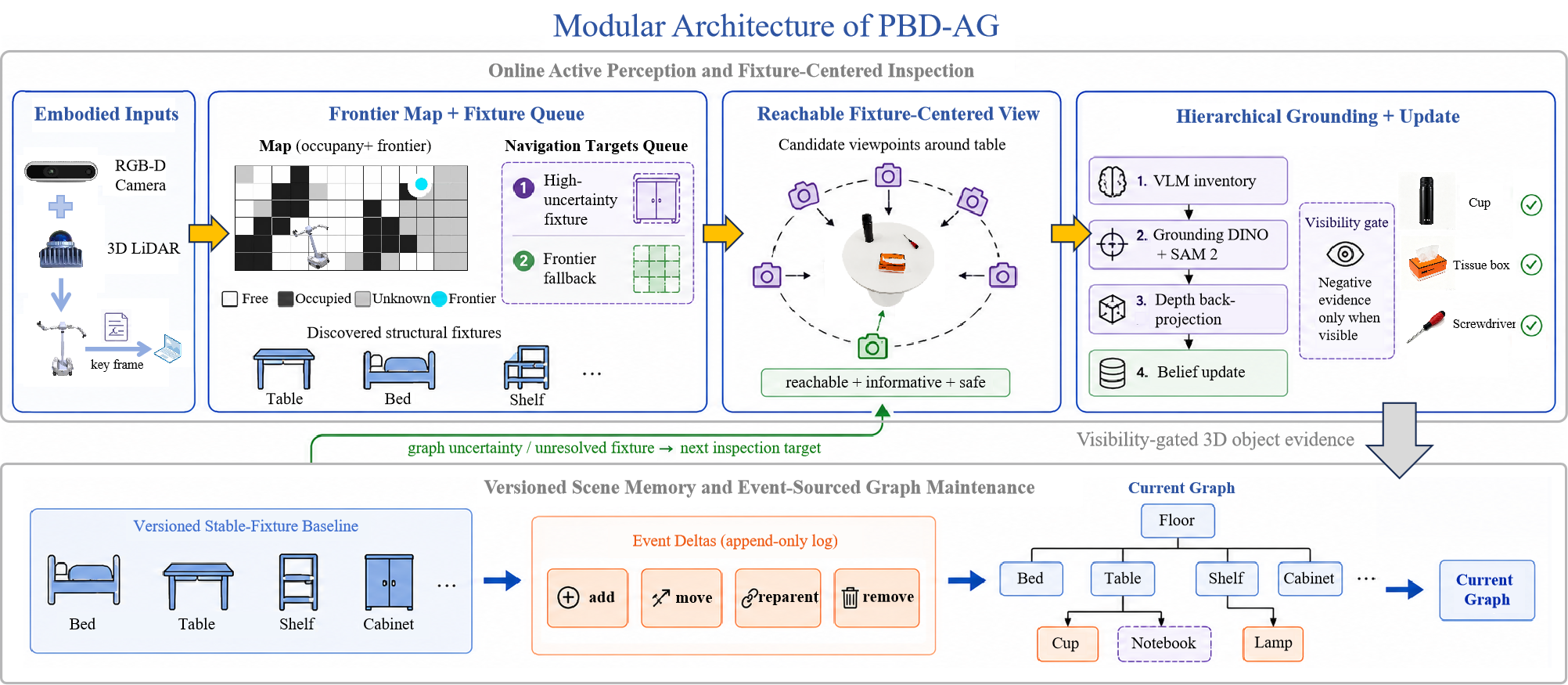}
    \caption{Architecture of PBD-AG. The robot maps unexplored space, confirms
coarse fixture tracks, and schedules task-eligible unresolved fixtures for
reachable close-view inspection. Confirmed fixtures populate the working
structural memory, which is frozen as an immutable baseline package;
subsequent lifecycle changes update persistent tracks while typed events
append an ordered audit trail.}
    \label{fig:architecture}
\end{figure*}

\section{Method}

\subsection{Problem Formulation}

At time $t$, the robot receives a posed RGB-D observation
$z_t=(I_t,D_t,K_t,T_t)$ together with range measurements used to maintain a
2D occupancy map $A_t$. Here, $K_t$ denotes the camera intrinsics, and $T_t$
transforms points from the camera frame into a shared world frame.

We maintain the online state and publish the current graph as
\begin{equation}
\begin{aligned}
S_t&=(\widetilde{B}_t,M_t,\mathcal{E}_{1:k_t},A_t),\\
G_t&=\operatorname{Publish}(\widetilde{B}_t,M_t),
\end{aligned}
\label{eq:state_publish}
\end{equation}
where $M_t$ is the persistent runtime track table,
$\widetilde{B}_t$ is its canonical-fixture view, and
$\mathcal{E}_{1:k_t}=(e_1,\ldots,e_{k_t})$ is the ordered audit-event stream.
Raw hypotheses $\mathcal{H}_t\subset M_t$ are unpublished tracks with
persistent identities. Promotion changes the lifecycle state of the same
track rather than creating a second node. Tracks in
$\mathcal{C}_t\subset M_t$ are authoritative confirmed fixtures represented
in $\widetilde{B}_t$ and published to $G_t$.

At episode finalization, the immutable structural package is
$B^{(0)}=\operatorname{Freeze}(\widetilde{B}_{t_b})$. Maintenance loads
$B^{(0)}$ into the runtime state without overwriting it. The post-bootstrap
mutable track state and its event suffix form the runtime delta maintained
alongside this baseline. Verified fixture-centered fine graphs attach to
canonical parent identities through the capture ledger.

Each persistent track $i$ in $M_t$ maintains
\begin{equation}
b_i^t=
\{
p_i^{\rm cls},
p_i^{\rm exist},
\mu_i,
\Sigma_i,
p_i^{\rm parent},
a_i,
\tau_i,
h_i,
s_i^{\rm life}
\},
\label{eq:belief}
\end{equation}
where component-level time indices are omitted for readability. The state
contains class and existence beliefs, a 3D center and recent-position
covariance, support-parent belief, an optional appearance embedding, the most recent
reliable observation time, a persistent identity $h_i$, and a lifecycle state
$s_i^{\rm life}$.

\subsection{Robot-Bootstrapped Baseline and Grounded Observations}
\label{sec:bootstrap}

During autonomous bootstrapping, PBD-AG initializes the working structural
memory as $\widetilde{B}_0=\varnothing$, while all cells in the occupancy map
$A_0$ are unknown. As frontier exploration proceeds, range sensing
incrementally expands the known region of $A_t$, and posed RGB-D observations
accumulate geometry and SigLIP-encoded open-vocabulary features
\citep{zhai2023siglip} in the shared world frame.

At each semantic update, the accumulated voxel map is queried against the
fixture taxonomy. Map points are retained when their class similarity exceeds
a class-specific threshold and has a sufficient margin over competing
classes. The retained points are clustered by DBSCAN in metric space, after
which geometrically implausible clusters are discarded. The resulting fixture
proposals are associated across semantic updates using class consistency,
center distance, and footprint overlap, forming persistent candidate tracks
$\mathcal{H}_t$.

A candidate must receive support from multiple temporally distinct observation
batches before publication. Stable, high-margin tracks passing the cross-frame
sensor-consensus gate are promoted directly to $\mathcal{C}_t$; near-threshold
or class-conflicting tracks receive one bounded, cached coarse-stage VLM
verification. Promotion publishes the same persistent identity $h_i$ as an
authoritative coarse node. Only canonical fixtures whose class belongs to the
task inspection taxonomy $\mathcal{T}_{\rm insp}$ and whose close-range fine
observation remains unresolved enter $\mathcal{Q}_t$; rejected hypotheses are
retired. Thus, not every structural fixture requires fine inspection.

After reaching a fixture-centered viewpoint, the fixture retains its canonical
identity $h_i$. Parent-fixture verification and fine-object grounding are then
performed as described in Section~3.4. Successful processing attaches verified
child nodes and support edges to the existing fixture node and marks its
close-range inspection complete.

At an accepted fixture-centered view, an independent VLM query first verifies
the target fixture class at its projected support region. Conditional on this
check, a VLM-based multimodal inventory proposes task-relevant fine categories
and counts. Grounding DINO \citep{liu2023groundingdino} localizes the selected
phrases, and SAM~2 \citep{ravi2024sam2} refines their masks. Valid mask depth
is back-projected through $(K_t,T_t)$; robust median and inlier filtering
suppress boundary leakage and produce proposal center $\mu_j$. SigLIP supplies
proposal-level appearance evidence.

Each accepted fine proposal records detector confidence, mask and valid-depth
support, proposal-level appearance, and its geometrically assigned parent.
Within the selected primary RGB-D view, non-maximum suppression, cross-class
overlap suppression, and support constraints form a local fine graph.
Alternate fixed-base camera angles are used for parent verification or bounded
observation retry when the primary image is unusable; they are not fused as
cross-view masks, point clouds, or 3D proposals. Successful processing attaches
verified child nodes and support edges to the existing canonical fixture and
marks its close-range inspection complete.

\subsection{Evidence-Gated Persistent Update}
\label{sec:persistent_update}

Coarse fixture clusters and current-frame dynamic proposals have different
geometric structure and failure modes. PBD-AG therefore uses type-specific
association front ends followed by a shared persistent lifecycle update.
Sensor outputs enter the update as explicit admission gates, association
tests, and fixed evidence increments rather than being collapsed into a
learned scalar reliability.

\paragraph{Type-specific association.}
Coarse fixture proposals are greedily associated using same-class planar
proximity, expanded-footprint coverage, and footprint IoU. A strictly
contained residual cluster may attach to an already matched fixture to absorb
partial-view fragmentation without creating another persistent identity.
Dynamic proposals require class agreement, a 3D center distance below
$0.8\,\mathrm{m}$, and SigLIP cosine similarity above $0.78$; the appearance
gate is omitted when an embedding is unavailable. Eligible pairs are greedily
matched one-to-one. A unique-class fallback permits persistent-ID recovery
after large motion when appearance does not conflict. Exact fixture gates and
fragment-attachment thresholds are reported in the supplementary material.

\paragraph{Persistent state fusion.}
For a matched pair $(i,j)$, accumulated class and parent evidence is normalized
into $p_i^{\rm cls}$ and $p_i^{\rm parent}$. Ordinary observations add unit
evidence, while cached VLM class verification contributes weight $3$.
Dynamic geometry uses the latest accepted RGB-D proposal; coarse fixture
centers use an exponential moving average with $\alpha=0.35$. The covariance
$\Sigma_i$ summarizes at most the 32 most recent accepted centers and is not
used for association. Appearance embeddings are averaged by observation count
and then $\ell_2$-normalized.

\begin{figure}[t]
    \centering
    \includegraphics[trim=0 2pt 6pt 0,clip,width=\columnwidth]
    {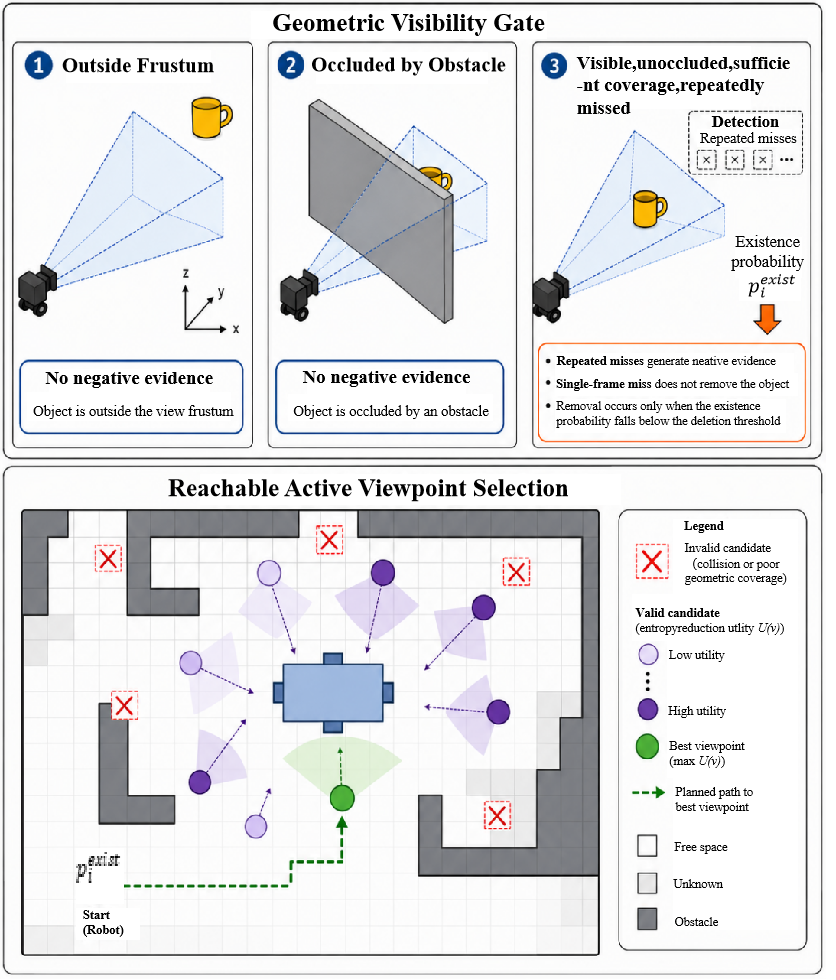}
    \caption{Visibility gating and graph-guided inspection.
Top: only geometrically observable misses reduce existence belief.
Bottom: graph uncertainty selects an unresolved fixture, after which feasible
views are ranked by coverage, standoff, travel, and obstacle clearance.}
    \label{fig:mechanisms}
\end{figure}

\paragraph{Visibility-gated existence.}
Here, $t$ indexes semantic update batches. Let
$\ell_i^t=\log[p_i^{\rm exist,t}/(1-p_i^{\rm exist,t})]$ denote the existence
log-odds. We update
\begin{equation}
\ell_i^t=\ell_i^{t-1}
+\eta_oO_i^t+\eta_cK_i^t
-\eta_mU_i^tV_i(t)-\eta_rR_i^t,
\label{eq:existence}
\end{equation}
where $O_i^t$, $K_i^t$, $U_i^t$, and $R_i^t$ denote an ordinary positive
observation, cached confirmation, unmatched track, and cached rejection. The
frozen increments are
$(\eta_o,\eta_c,\eta_m,\eta_r)=(1.0,1.8,0.70,2.20)$.

Visibility is evaluated by projecting samples from the stored world-frame 3D
bounding box. A sample must lie in the valid camera range, remain inside the
image boundary, and have valid depth in a $5\times5$ window. It is considered
unoccluded when
$d_{\rm meas}+0.15\,\mathrm{m}\ge d_{\rm expected}$, and $V_i(t)=1$ only when
at least two samples pass. Hence an unmatched but out-of-view or occluded track
receives no negative evidence, and neither its log-odds nor visible-miss count
changes.

\paragraph{Lifecycle transitions and audit events.}
Creating an internal hypothesis emits \texttt{add}, while graph publication
emits \texttt{confirm} and requires cross-batch confirmation with
$p_i^{\rm exist}\ge0.72$. A positive matched batch emits \texttt{update};
motion of at least $0.35\,\mathrm{m}$ emits \texttt{move}, and a valid support
change emits \texttt{reparent}. A published dynamic node is
\texttt{demote}d when $p_i^{\rm exist}<0.30$ and emits \texttt{remove} only
when $p_i^{\rm exist}\le0.12$ after at least three visibility-admissible
misses. Pinned fixtures are exempt from dynamic removal. Removed tracks remain
in memory and recover their original identity upon successful reassociation.

Each audit event stores its total-order sequence, frame, persistent identity,
lifecycle state, and transition details. The maintained structural and track
states publish the current graph, while the event stream records update
provenance. Fine nodes from active inspection are attached through the capture
ledger and do not re-enter the frame-level dynamic association front end.

\subsection{Graph-Uncertainty-Guided Fixture-Centered Inspection}

Let $\mathcal{Q}_t\subseteq\mathcal{C}_t$ contain canonical confirmed fixtures
whose close-range inspection remains unresolved. Completed, rejected,
abandoned, currently processed, and retry-cooldown tracks are excluded.

We represent fixture uncertainty as
\begin{equation}
u_i^t=\mathbf{w}^{\top}\boldsymbol{\phi}_i^t,\qquad
\mathbf{w}=(.20,.30,.10,.15,.10,.15),
\label{eq:fixture_uncertainty}
\end{equation}
where $\boldsymbol{\phi}_i^t$ contains normalized existence, class, parent,
position, age, and confirmation uncertainties, all increasing from $0$ to $1$.
Existence ambiguity is $4p_i^{\rm exist}(1-p_i^{\rm exist})$, while class
uncertainty combines normalized class entropy and score-margin ambiguity.
All transforms and weights are fixed before evaluation.

Among fixtures with at least one feasible view, the scheduler selects
\begin{equation}
i_t^*=
\arg\max_{i\in\mathcal{Q}_t^{\rm feas}}
\left[0.8u_i^t+0.2\operatorname{Prox}(i)\right],
\label{eq:inspection_target}
\end{equation}
where $\operatorname{Prox}(i)$ is a clipped proximity score that decreases to
zero over $8\,\mathrm{m}$. The selected target remains locked until capture,
failure, rejection, or completion.

Candidate views are connected known-free cells satisfying standoff,
inflated-map reachability, line-of-sight, projected-coverage, and image-margin
constraints. Among feasible views, the controller minimizes
\begin{equation}
\begin{aligned}
J(v,i)=&
|d_{\rm foot}(v,i)-d_{\rm pref}|
+0.08D_t(v)-0.20C(v)\\
&+4[1-\operatorname{Cov}(v,i)]
+P_{\rm margin}(v,i),
\end{aligned}
\label{eq:inspection_view}
\end{equation}
where $d_{\rm pref}=0.90\,\mathrm{m}$, $D_t(v)$ is robot-to-view distance,
$C(v)=\min(\operatorname{Clr}(v),1\,\mathrm{m})$, and
$\operatorname{Cov}$ is predicted support-surface coverage. Previously failed
views are excluded.

If a fixture has no feasible view, it is deferred while the controller checks
the next queued target; frontier exploration resumes when none is reachable.
At the selected pose, a fixed-base RGB-D sweep provides one contract-valid
primary view for fine-object grounding, while alternate angles support parent
verification and retry rather than cross-view proposal fusion. Successful
processing attaches verified child nodes and support edges to the canonical
fixture identity; failures retain the target for bounded retry. Thus geometric
frontiers expand unknown space, whereas graph uncertainty schedules active
inspection of already-discovered fixtures.

\section{Experiments}


\begin{table*}[t]
\centering
\small
\setlength{\tabcolsep}{6.5pt}
\begin{tabular}{lccccc}
\toprule
Method & P$\uparrow$ & R$\uparrow$ & F1$\uparrow$ &
Err.\,[m]$\downarrow$ & MAE$\downarrow$\\
\midrule
\textbf{PBD-AG}
& \textbf{.951$\pm$.091} & .802$\pm$.122
& \textbf{.868$\pm$.102} & .320$\pm$.119
& \textbf{.356$\pm$.233}\\
DynaMem-adapted
& .890$\pm$.156 & .667$\pm$.175
& .757$\pm$.163 & \textbf{.318$\pm$.127}
& .716$\pm$.389\\
ConceptGraphs-adapted
& .598$\pm$.081 & \textbf{.838$\pm$.136}
& .696$\pm$.095 & .323$\pm$.121
& 1.176$\pm$.501\\
\bottomrule
\end{tabular}
\caption{Controlled coarse-fixture comparison under shared evidence.
All methods receive identical proposal-derived fixture streams. Values are
mean$\pm$sample standard deviation over nine scene--seed runs.}
\label{tab:shared_replay}
\end{table*}

\begin{table*}[t]
\centering
\footnotesize
\setlength{\tabcolsep}{5.5pt}

\begin{tabular}{lccccc}
\toprule
Method & IDF1$\uparrow$ & ID-Pres.\,$\uparrow$ & IDS$\downarrow$ &
Event R.$\uparrow$ & False abs.$\downarrow$\\
\midrule
\textbf{PBD-AG}
& \textbf{.833$\pm$.029} & .845$\pm$.034 & \textbf{.0$\pm$.0}
& \textbf{11/12} & .014\\
\midrule
\multicolumn{6}{l}{\emph{Controlled memory baselines}}\\[-1pt]
DynaMem-inspired
& .688$\pm$.025 & \textbf{.938$\pm$.057} & 1.0$\pm$.0
& 8/12 & .034\\
ConceptGraphs-inspired
& .615$\pm$.024 & .899$\pm$.000 & 1.0$\pm$.0
& 6/12 & .017\\
Last observation
& .702$\pm$.026 & \textbf{.938$\pm$.057} & 1.0$\pm$.0
& 8/12 & .017\\
\midrule
\multicolumn{6}{l}{\emph{Alternative update rules}}\\[-1pt]
Append-only
& .627$\pm$.025 & .899$\pm$.000 & 1.0$\pm$.0
& 6/12 & .000$^\dagger$\\
Static-only
& .146$\pm$.000 & .179$\pm$.000 & .0$\pm$.0
& 0/12 & .000$^\dagger$\\
Online remap
& .083$\pm$.003 & .125$\pm$.000 & 17.67$\pm$.58
& 6/12 & .626\\
\midrule
\multicolumn{6}{l}{\emph{Component ablations}}\\[-1pt]
w/o visibility gate
& .708$\pm$.029 & .741$\pm$.008 & .0$\pm$.0
& 6/12 & .262\\
w/o persistent ID
& .070$\pm$.003 & .125$\pm$.000 & 43.33$\pm$1.15
& 6/12 & .014\\
\bottomrule
\end{tabular}

\caption{Controlled dynamic-memory evaluation on shared grounded 3D streams.
Values are mean$\pm$sample standard deviation over three seeds.
$^\dagger$ indicates zero false absence because disappearance is never
resolved.}
\label{tab:dynamic_memory}
\end{table*}

We evaluate whether PBD-AG (Q1) autonomously constructs hierarchical scene
graphs in initially unknown environments, (Q2) improves coarse association and
retention under shared evidence, and (Q3) preserves identity and recovers
task-relevant changes over time.

\subsection{Experimental Setup}

\paragraph{Autonomous construction protocol.}
We conduct nine integrated rollouts in OmniGibson/BEHAVIOR
\citep{li2024behavior}, using Gates bedroom, Hotel suite, and Benevolence with
seeds 7, 17, and 27. Every rollout begins with empty working structural memory
and an occupancy map whose cells are initially unknown. The coarse-discovery
budget is 600 semantic keyframes. Once exhausted, no new coarse candidate is
admitted, but already queued fine inspections may complete before terminal
evaluation.

Ground-truth identities, semantic annotations, and correspondences are never
available to the system. Simulator camera poses are used only for RGB-D
registration, and an OmniGibson shortest-path proxy executes navigation goals
selected by the system. Frontier and fixture-inspection scheduling do not use
ground-truth scene annotations. A single scene-level taxonomy and
parameterization is held fixed across the three reported seeds, with no
seed-specific setting. Development and run-selection provenance, resolved
configurations, and artifact hashes are provided in the supplementary
material.

\paragraph{Metrics and controlled comparisons.}
Coarse predictions are greedily matched one-to-one to same-class ground-truth
instances within $1.0\,\mathrm{m}$. Fine matching additionally requires the
correct matched fixture parent and center distance below $0.5\,\mathrm{m}$;
its F1 therefore evaluates both object recovery and hierarchical grounding.
We report instance precision, recall, F1, and matched-center error. Count MAE
averages class-wise instance-count error. Unless identified as pooled,
variation is the sample standard deviation over seeds.

For controlled coarse comparison, PBD-AG and association/update controls
adapted from DynaMem \citep{liu2024dynamem} and ConceptGraphs
\citep{gu2024conceptgraphs} consume the same selected episodes and
proposal-derived fixture observations. \emph{Adapted} denotes a
common-interface implementation isolating graph association and memory update
from detector, sensing, and navigation differences; it is not an end-to-end
reproduction of either original system. Evaluation is restricted to the
category intersection supported by this interface and is reported separately
from autonomous full-graph evaluation.

The shared-evidence subset may therefore be smaller than the full-scene
ground truth in Table~\ref{tab:end_to_end}. In the tables, Err.\ denotes
matched-center error and MAE denotes class-wise count error. For the dynamic
protocol, event recall is pooled over scheduled events and false absence over
eligible post-acquisition checkpoints.

\subsection{Autonomous Scene Construction and Coarse Comparison}

\begin{table*}[t]
\centering
\small
\setlength{\tabcolsep}{3.5pt}

\begin{tabular}{lccccccccc}
\toprule
& \multicolumn{4}{c}{Coarse fixtures} &
\multicolumn{4}{c}{Fine-grained objects} \\
\cmidrule(lr){2-5}\cmidrule(lr){6-9}
Scene (coarse/fine GT) &
P$\uparrow$ & R$\uparrow$ & F1$\uparrow$ & Err.\,[m]$\downarrow$ &
P$\uparrow$ & R$\uparrow$ & F1$\uparrow$ & Err.\,[m]$\downarrow$ \\
\midrule
Gates bedroom (9/15)
& 1.000$\pm$.000 & .926$\pm$.064 & .961$\pm$.034 & .215$\pm$.035
& .874$\pm$.050 & .889$\pm$.077 & .878$\pm$.023 & .036$\pm$.009 \\
Hotel suite (13/14)
& .944$\pm$.048 & .795$\pm$.089 & .859$\pm$.036 & .320$\pm$.029
& .963$\pm$.064 & .571$\pm$.143 & .710$\pm$.117 & .071$\pm$.012 \\
Benevolence (14/12)
& .854$\pm$.109 & .714$\pm$.124 & .777$\pm$.119 & .464$\pm$.043
& .617$\pm$.114 & .500$\pm$.144 & .550$\pm$.128 & .029$\pm$.002 \\
\midrule
All rollouts
& .933$\pm$.088 & .812$\pm$.124 & .866$\pm$.102 & .333$\pm$.113
& .818$\pm$.170 & .653$\pm$.210 & .713$\pm$.167 & .045$\pm$.021 \\
\bottomrule
\end{tabular}

\caption{Autonomous PBD-AG scene construction.
Mean$\pm$sample standard deviation over three seeds per scene. The final row
aggregates all nine rollouts; values in parentheses are the fixed
coarse/fine ground-truth instance counts.}
\label{tab:end_to_end}
\end{table*}

Across the nine autonomous rollouts, PBD-AG attains coarse and fine F1 of
$0.866\pm0.102$ and $0.713\pm0.167$, respectively
(Table~\ref{tab:end_to_end}). Gates yields the strongest joint recovery, with
$0.961$ coarse and $0.878$ fine F1. Hotel retains high fine-grained precision
($0.963$) but lower recall, indicating that its fine errors are dominated by
missed objects rather than spurious insertions. Performance decreases in
Benevolence, where repeated fixture geometry and restricted viewpoints
increase cross-instance ambiguity and delineate the current operating regime
of the system.

\begin{figure}[!htbp]
    \centering
    \includegraphics[width=0.8\columnwidth]{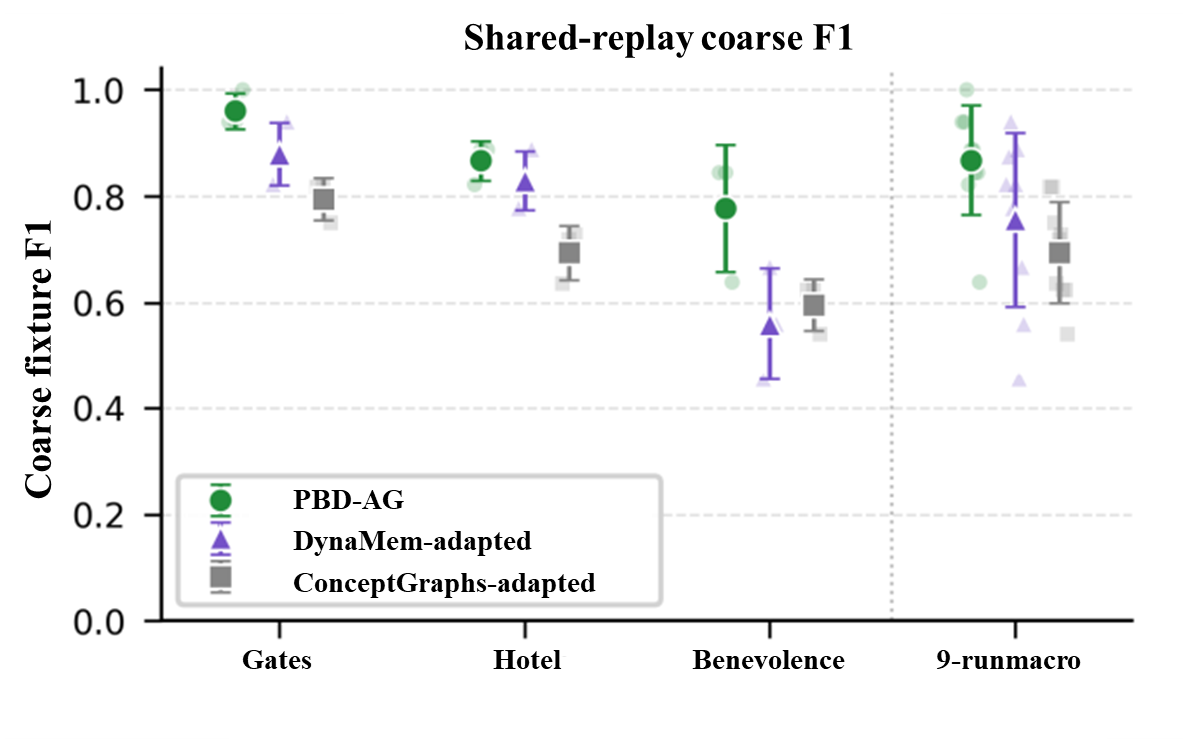}
    \caption{Autonomous fixture recovery across environments.
    Scene-wise coarse-fixture F1 under shared evidence; PBD-AG attains the
    highest mean F1 in all three environments.}
    \label{fig:coarse_recovery_single}
\end{figure}

Under shared evidence, PBD-AG obtains the highest nine-run fixture F1
($0.868\pm0.102$), exceeding DynaMem-adapted and ConceptGraphs-adapted by
11.1 and 17.2 points, respectively, while reducing count MAE from $0.716$ to
$0.356$ relative to the stronger adapted control
(Table~\ref{tab:shared_replay}). The scene-wise breakdown in
Figure~\ref{fig:coarse_recovery_single} shows that its mean F1 is higher in
all three environments. Matched-center errors remain similar because the
methods receive common geometric evidence; the principal differences arise
from instance association, consolidation, and retention.
\begin{figure*}[!t]
    \centering
    \includegraphics[width=0.97\textwidth]{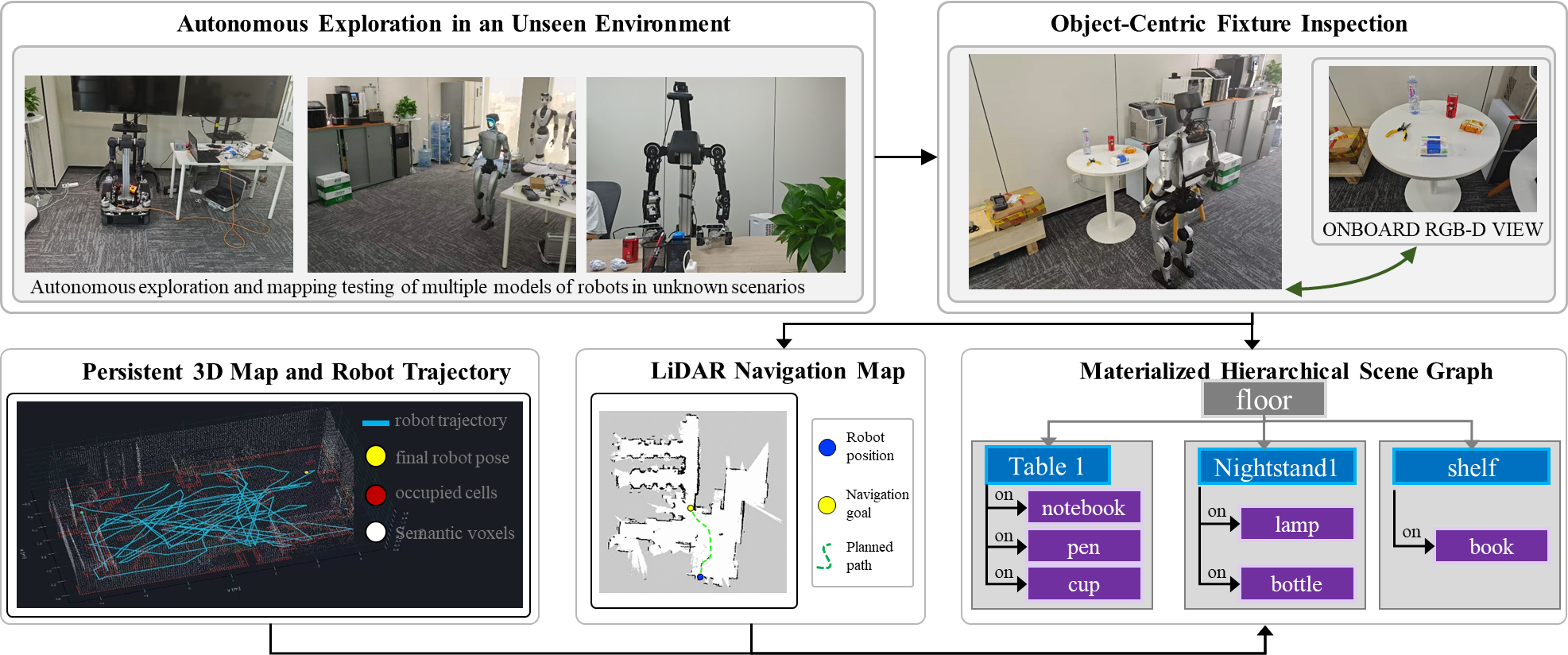}
    \caption{Qualitative physical-robot deployment.
    \textbf{(A)} Autonomous exploration; \textbf{(B)} object-centric RGB-D
    inspection; \textbf{(C)} persistent 3D mapping with robot trajectory and
    inspection goals; \textbf{(D)} LiDAR occupancy mapping; and
    \textbf{(E)} the materialized hierarchical scene graph.}
    \label{fig:physical_demo}
\end{figure*}
\subsection{Dynamic Persistence and Component Analysis}

\paragraph{Controlled dynamic protocol.}
We evaluate persistent memory on three Hotel sequences of 300 frames. Each
seed contains one appearance, within-support motion, disappearance, and
cross-support transfer, giving 12 scheduled events. Within each seed, every
method receives the same grounded 3D stream, while event times and identities
remain hidden. We report IDF1 \citep{ristani2016performance}, presence F1 after
global identity assignment (ID-Pres.), identity switches (IDS), pooled event
recall, and post-acquisition false absence over ground-truth-present eligible
checkpoints. \emph{Inspired} rows instantiate only the corresponding temporal
update rule; the remaining controls isolate alternative rules or individual
PBD-AG components.

\begin{figure}[t]
    \centering
    \includegraphics[width=0.8\columnwidth]{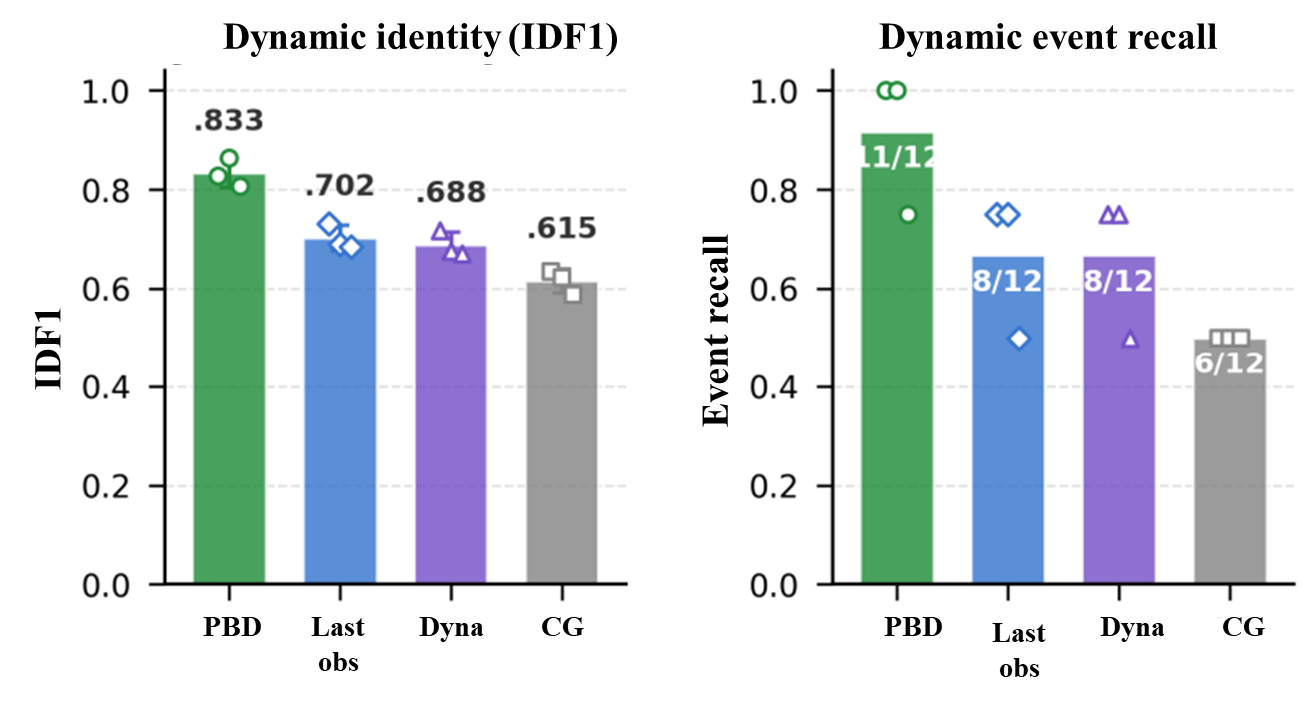}
    \caption{Dynamic identity and event recall.
    IDF1 and pooled event recall on the shared 3D observation streams.}
    \label{fig:dynamic_memory_single}
\end{figure}

\paragraph{Persistent-memory results.}
PBD-AG achieves the highest IDF1 ($0.833\pm0.029$), recovers 11 of 12
scheduled events, and incurs no identity switch
(Table~\ref{tab:dynamic_memory} and
Figure~\ref{fig:dynamic_memory_single}). Immediate-update controls
attain higher presence F1 by publishing single observations without
cross-frame confirmation. PBD-AG instead trades immediate publication for
identity and event continuity, improving IDF1 over the strongest such control
by 13.1 points and event recall from 8/12 to 11/12.

\paragraph{Component ablations.}
Without visibility gating, false absence rises from $0.014$ to $0.262$ and
event recall falls from 11/12 to 6/12. Without persistent IDs, IDF1 falls from
$0.833$ to $0.070$ and IDS rises to $43.33\pm1.15$. Online remapping also
loses continuity; append-only and static-only avoid false absence only by
failing to resolve disappearance. These results isolate the contributions of
geometric visibility gating and persistent identity.


\subsection{Qualitative Physical-Robot Demonstration}

We deploy the exploration-to-graph pipeline on physical mobile robots without
a preloaded scene graph. Onboard RGB-D and LiDAR observations build working
structural memory, support fixture-centered inspection, and publish a
hierarchical graph while preserving canonical fixture identities.
Figure~\ref{fig:physical_demo} shows a representative execution, including
autonomous exploration, close-range inspection, registered 3D and occupancy
maps, and the published graph.

The supplementary video provides the complete temporal rollout together with
additional qualitative deployments on multiple robot platforms in previously
unseen indoor environments. These demonstrations assess system integration
and transfer qualitatively and are not included in the quantitative
comparisons above.

\section{Conclusion}

We presented PBD-AG, an autonomous baseline--delta scene graph for persistent robot memory. PBD-AG bootstraps stable fixtures through onboard exploration, attaches dynamic objects to canonical fixtures, and revises their states using geometrically admissible evidence. Across multi-scene and controlled dynamic-memory evaluations, PBD-AG achieves higher coarse-fixture F1 than adapted controls, preserves identity without switches, and detects 11 of 12 object-state events. Paired acquisition experiments further support graph-conditioned fixture prioritization under bounded budgets. Physical-robot deployment demonstrates integrated baseline construction, grounded inspection, and event-traced updates, yielding a compact, traceable world model for downstream robotic systems.

\bibliography{aaai2027}
\end{document}